\pdfoutput=1

\documentclass[11pt]{article}

\usepackage[]{EMNLP2022}

\usepackage{times}
\usepackage{latexsym}
\usepackage{enumitem}
\usepackage{multirow}
\usepackage{multicol}
\usepackage{amsmath}
\usepackage{xcolor}
\usepackage{graphicx}
\usepackage{array}
\usepackage{booktabs}
\usepackage{subcaption}
\usepackage{soul}
\usepackage{enumitem}

\usepackage{algorithm}
\usepackage[noend]{algpseudocode}
\algrenewcommand\algorithmicrequire{\textbf{Input:}}
\algrenewcommand\algorithmicensure{\textbf{Output:}}

\newcommand{\zblue}[1]{\text{\textcolor{blue}{#1}}}

\usepackage[T1]{fontenc}

\usepackage[utf8]{inputenc}

\usepackage{microtype}

\title{A Survey of Active Learning for Natural Language Processing}

\author{
Zhisong Zhang, Emma Strubell, Eduard Hovy\\
Language Technologies Institute, Carnegie Mellon University\\
\texttt{zhisongz@cs.cmu.edu, strubell@cmu.edu, hovy@cmu.edu}
}

\begin{document}
\maketitle

\begin{abstract}
In this work, we provide a literature review of active learning (AL) for its applications in natural language processing (NLP). In addition to a fine-grained categorization of query strategies, we also investigate several other important aspects of applying AL to NLP problems. These include AL for structured prediction tasks, annotation cost, model learning (especially with deep neural models), and starting and stopping AL. Finally, we conclude with a discussion of related topics and future directions.
\end{abstract}

\section{Introduction}
\label{sec:c5a:intro}

The majority of modern natural language processing (NLP) systems are based on data-driven machine learning models. The success of these models depends on the quality and quantity of the available target training data. While these models can obtain impressive performance if given enough supervision, it is usually expensive to collect large amounts of annotations, especially considering that the labeling process can be laborious and challenging for NLP tasks (\S\ref{sec:c5a:cost}). \emph{Active learning} (AL), an approach that aims to achieve high accuracy with fewer training labels by allowing a model to choose the data to be annotated and used for learning, is a widely-studied approach to tackle this labeling bottleneck \citep{settles2009active}. 

Active learning has been studied for more than twenty years \citep{lewis1994sequential,lewis1994heterogeneous,cohn1994improving,cohn1996active} and there have been several literature surveys on this topic \citep{settles2009active,olsson2009literature,fu2013survey,aggarwal2014active,hino2020active,schroder2020survey,ren2021survey,zhan2022comparative}.
Nevertheless, there is still a lack of an AL survey for NLP that includes recent advances. \citet{settles2009active} and \citet{olsson2009literature} provide great surveys covering AL for NLP, but these surveys are now more than a decade old. In the meantime, the field of NLP has been transformed by deep learning.
While other more recent surveys cover deep active learning, they are either too specific, focused only on text classification \citep{schroder2020survey}, or too general, covering AI applications more broadly \citep{ren2021survey,zhan2022comparative}.
Moreover, applying AL to NLP tasks requires specific considerations, e.g. handling complex output structures and trade-offs in text annotation cost (\S\ref{sec:c5a:A}), which have not been thoroughly discussed.

\begin{figure}[t]
	\centering
	\includegraphics[width=0.42\textwidth]{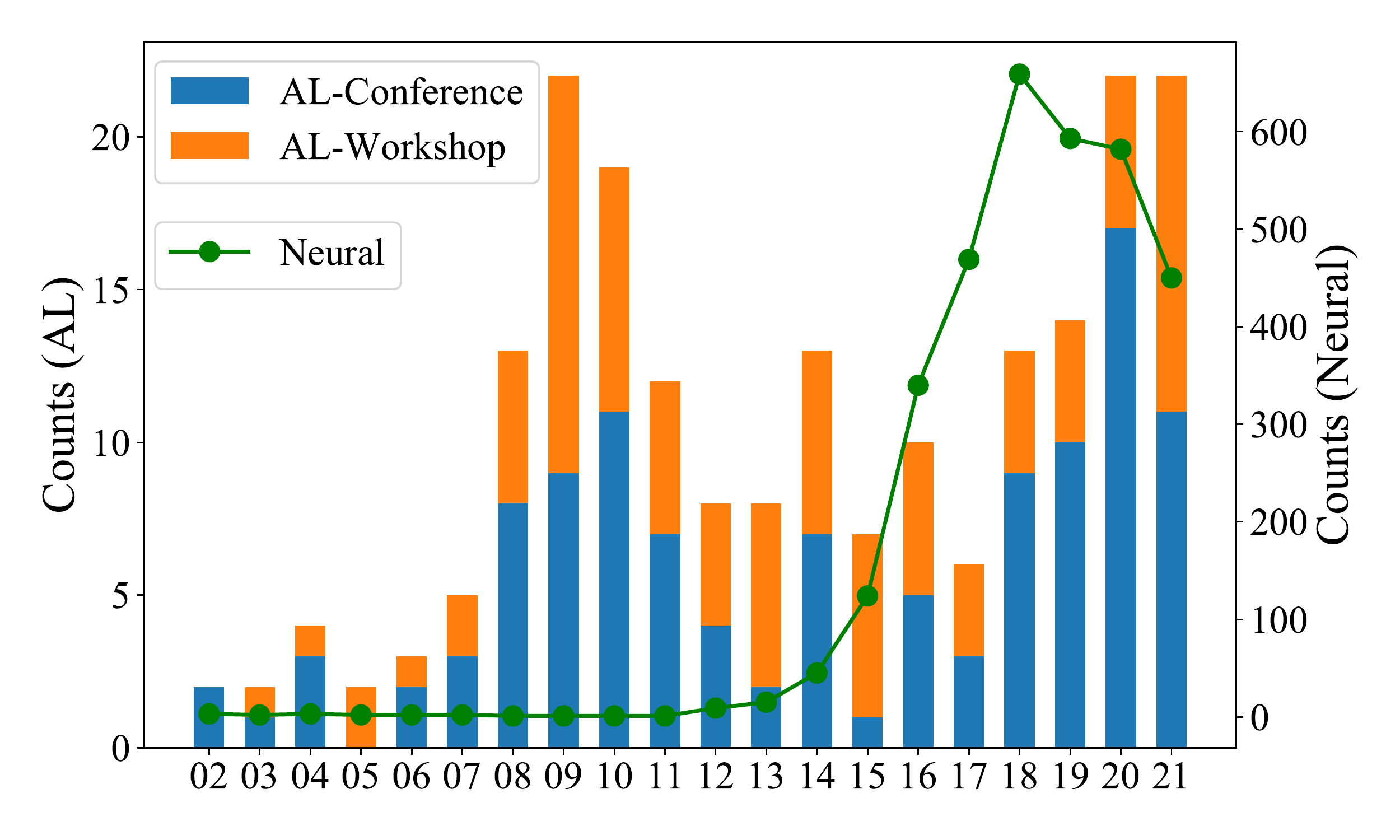}
	\vspace*{-4mm}
	\caption{Counts of AL (left) and \textit{``neural''} (right) papers in the ACL Anthology over the past twenty years.}
	\label{fig:c5a:counts}
	\vspace*{-4mm}
\end{figure}

In order to provide an NLP-specific AL survey,\footnote{The descriptions in this survey are mostly brief to provide more comprehensive coverage in a compact way. We hope that this work can serve as an index for corresponding works.} we start by searching the ACL Anthology for AL-related papers. We simply search for the keyword \textit{``active''} in paper titles and then perform manual filtering. We also gradually include relevant papers missed by keyword search and papers from other venues encountered by following reference links throughout the surveying process.\footnote{Appendix~\ref{sec:c5a:pt} describes more details of the process.} The distribution of AL-related papers in the ACL Anthology over the past twenty years is shown in Figure~\ref{fig:c5a:counts}, which also includes rough counts of works concerning neural models by searching for the word \textit{``neural''} in titles. The overall trend is interesting. There is a peak around the years of 2009 and 2010, while the counts drop and fluctuate during the mid-2010s, which corresponds to the time when neural models became prominent in NLP. We observe a renewed interest in AL research in recent years, which is primarily focused on deep active learning \citep{ren2021survey,zhan2022comparative}.


\subsection{Overview}
\label{sec:c5a:over}


\begin{algorithm}[t]
    \caption{A typical active learning procedure.}
	\label{algo:c5a:al}
	\begin{algorithmic}[1]
		\small
		\Require An unlabeled data pool $\mathcal{U}$.
		\Ensure The final labeled dataset $\mathcal{L}$ and trained model $\mathcal{M}$.
		\State $\mathcal{L},~\mathcal{U}~\leftarrow~\text{seed}(\mathcal{U})$ \Comment{\zblue{\textbf{Start} (\S\ref{sec:c5a:start})}}
		\State $\mathcal{M}~\leftarrow~\text{train}(\mathcal{L},~\mathcal{U})$ \Comment{\zblue{\textbf{Model Learning} (\S\ref{sec:c5a:M})}}
		\While{\textbf{not} stop\_criterion()} \Comment{\zblue{\textbf{Stop} (\S\ref{sec:c5a:stop})}}
		\State $\mathcal{I}~\leftarrow~\text{query}(\mathcal{M}, ~\mathcal{U})$ \Comment{\zblue{\textbf{Query} (\S\ref{sec:c5a:Q}, \S\ref{sec:c5a:A})}}
		\State $\mathcal{I}'~\leftarrow~\text{annotate}(\mathcal{I})$ \Comment{\zblue{\textbf{Annotate} (\S\ref{sec:c5a:A})}}
		\State $\mathcal{U}~\leftarrow~\mathcal{U} - \mathcal{I};~~\mathcal{L}~\leftarrow~\mathcal{L} \cup \mathcal{I}'$
		\State $\mathcal{M}~\leftarrow~\text{train}(\mathcal{L},~\mathcal{U})$ \Comment{\zblue{\textbf{Model Learning} (\S\ref{sec:c5a:M})}}
		\EndWhile
		\State \textbf{return} $\mathcal{L},~\mathcal{M}_f$
	\end{algorithmic}
\end{algorithm}

We mainly examine the widely utilized pool-based scenario \citep{lewis1994sequential}, where a pool of unlabeled data is available and instances are drawn from the pool to be annotated. Algorithm~\ref{algo:c5a:al} illustrates a typical AL procedure, which consists of a loop of instance selection with the current model and model training with updated annotations.
The remainder of this survey is organized corresponding to the main steps in this procedure:
\begin{itemize}[noitemsep,leftmargin=*,topsep=0.5mm]
\item In \S\ref{sec:c5a:Q}, we discuss the core aspect of AL: Query strategies, with a fine-grained categorization over informativeness (\S\ref{sec:c5a:Qi}), representativeness (\S\ref{sec:c5a:Qr}) and the combination of these two (\S\ref{sec:c5a:Qh}).
\item In \S\ref{sec:c5a:A}, we cover the two additional important topics of querying and annotating for NLP tasks: AL for structured prediction tasks (\S\ref{sec:c5a:sp}) and the cost of annotation with AL (\S\ref{sec:c5a:cost}).
\item In \S\ref{sec:c5a:M}, we discuss model and learning: the query-successor model mismatch scenario (\S\ref{sec:c5a:mismatch}) and AL with advanced learning techniques (\S\ref{sec:c5a:learn}).
\item In \S\ref{sec:c5a:S}, we examine methods for starting (\S\ref{sec:c5a:start}) and stopping (\S\ref{sec:c5a:stop}) AL.
\end{itemize}
In \S\ref{sec:c5a:O}, some other aspects of AL for NLP are discussed. In \S\ref{sec:c5a:dir}, we conclude with related and future directions. We also include representative AL works for various NLP tasks in Appendix~\ref{sec:c5a:T}.


\section{Query Strategies}
\label{sec:c5a:Q}

\subsection{Informativeness}
\label{sec:c5a:Qi}

Informativeness-based query strategies mostly assign an informative measure to each unlabeled instance \emph{individually}. The instance(s) with the highest measure will be selected.

\subsubsection{Output Uncertainty}
\label{sec:c5a:Quncer}

\textbf{Uncertainty sampling} \citep{lewis1994sequential} is probably the simplest and the most commonly utilized query strategy. It prefers the most uncertain instances judged by the model outputs. For probabilistic models, entropy-based \citep{shannon1948mathematical}, least-confidence \citep{culotta2005reducing} and margin-sampling \citep{scheffer2001active,schein2007active} are three typical uncertainty sampling strategies \citep{settles2009active}.
\citet{schroder-etal-2022-revisiting} revisit some of these uncertainty-based strategies with Transformer-based models and provide empirical results for text classification.
For non-probabilistic models, similar ideas can be utilized, such as selecting the instances that are close to the decision boundary in an SVM \citep{schohn2000less,tong2001support}.

Another way to measure output uncertainty is to check the divergence of a model's predictions with respect to an instance's local region. If an instance is near the decision boundary, the model's outputs may be different within its local region. In this spirit, recent works examine different ways to check instances' \textbf{local divergence}, such as nearest-neighbour searches \citep{margatina-etal-2021-active}, adversarial perturbation \citep{zhang-etal-2022-allsh} and data augmentation \citep{jiang-etal-2020-camouflaged}.


\subsubsection{Disagreement}
\label{sec:c5a:Qcom}

Uncertainty sampling usually considers the outputs of only one model. In contrast, \textbf{disagreement-based strategies} utilize multiple models and select the instances that are most disagreed among them. This is also a widely-adopted algorithm, of which the famous query-by-committee \citep[QBC;][]{seung1992query} is an example. The disagreement can be measured by vote entropy \citep{engelson-dagan-1996-minimizing}, KL-divergence \citep{mccallum1998employing} or variation ratio \citep{freeman1965elementary}.

To construct the model committee, one can train a group of distinct models. Moreover, taking a Bayesian perspective over the model parameters is also applicable \citep{houlsby2011bayesian}. Especially with neural models, \citep{gal2016dropout} show that dropout could approximate inference and measure model uncertainty.
This \textbf{deep Bayesian method} has been applied to AL for computer vision (CV) tasks \citep{gal2017deep} as well as various NLP tasks \citep{siddhant-lipton-2018-deep,shen2018deep,shelmanov-etal-2021-active}.


\subsubsection{Gradient}
\label{sec:c5a:Qgrad}

\textbf{Gradient information} can be another signal for querying, with the motivation to choose the instances that would most strongly impact the model. In this strategy, informativeness is usually measured by the norm of the gradients. Since we do not know the gold labels for unlabeled instances, the loss is usually calculated as the expectation over all labels. This leads to the strategy of \textbf{expected gradient length} (EGL), introduced by \citet{settles2007multiple} and later applied to sequence labeling \citep{settles-craven-2008-analysis} and speech recognition \citep{huang2016active}. \citet{zhang2017active} explore a variation for neural networks where only the gradients of word embeddings are considered and show its effectiveness for text classification.


\subsubsection{Performance Prediction}
\label{sec:c5a:Qpred}

Predicting performance can be another indicator for querying.
Ideally, the selected instances should be the ones that most \textbf{reduce future errors} if labeled and added to the training set. This motivates the \textbf{expected loss reduction} (ELR) strategy \citep{roy2001toward}, which chooses instances that lead to the least expected error if added to retrain a model. This strategy can be computationally costly since retraining is needed for each candidate. Along this direction, there have been several recent updates on the estimation metrics, for example, based on predicted error decay on groups \cite{chang2020using}, mean objective cost of uncertainty \citep{zhao2021uncertaintyaware} and strictly proper scoring rules \citep{tan2021diversity}.

Recently, methods have been proposed to learn another model to select instances that lead to the fewest errors, usually measured on a held-out development set. Reinforcement learning and imitation learning have been utilized to obtain the selection policy \citep{bachman2017learning,fang-etal-2017-learning,liu-etal-2018-learning-actively,liu-etal-2018-learning,zhang-etal-2022-active}.
This \textbf{learning-to-select strategy} may have some constraints. First, it requires labeled data (maybe from another domain) to train the policy. To mitigate this reliance, \citet{vu-etal-2019-learning} use the current task model as an imperfect annotator for AL simulations. Moreover, the learning signals may be unstable for complex tasks, as \citet{koshorek-etal-2019-limits} show for semantic tasks.

A similar and simpler idea is to select the most erroneous or ambiguous instances with regard to the current task model, which can also be done with another \textbf{performance-prediction model}. 
\citet{yoo2019learning} directly train a smaller model to predict the instance losses for CV tasks, which have been also adopted for NLP \citep{cai2021ne,shen2021active}.
In a similar spirit, \citet{wang2017active} employ a neural model to judge the correctness of the model prediction for SRL and \citet{brantley-etal-2020-active} learn a policy to decide whether expert querying is required for each state in sequence labeling.
Inspired by data maps \citep{swayamdipta-etal-2020-dataset}, \citet{zhang-plank-2021-cartography-active} train a model to select ambiguous instances whose average correctness over the training iterations is close to a pre-defined threshold.
For machine translation (MT), special techniques can be utilized to seek erroneous instances, such as using a backward translator to check round-trip translations \citep{haffari-etal-2009-active,zeng-etal-2019-empirical} or quality estimation \citep{logacheva-specia-2014-confidence,logacheva-specia-2014-quality,chimoto-bassett-2022-comet}.




\subsection{Representativeness}
\label{sec:c5a:Qr}

Only considering the informativeness of individual instances may have the drawback of sampling bias \cite{dasgupta2011two,prabhu-etal-2019-sampling} and the selection of outliers \citep{roy2001toward,karamcheti-etal-2021-mind}. Therefore, representativeness, which measures how instances correlate with each other, is another major factor to consider when designing AL query strategies.

\subsubsection{Density}
\label{sec:c5a:Qden}

With the motivation to avoid outliers, density-based strategies prefer instances that are more \textbf{representative of the unlabeled set}.
Selecting by $n$-gram or word counts \citep{ambati-etal-2010-active,zhao-etal-2020-active} can be regarded as a simple way of density measurement.
Generally, the common measurement is an instance's average similarity to all other instances \citep{mccallum1998employing,settles-craven-2008-analysis}. While it may be costly to calculate similarities of all instance pairs, considering only $k$-nearest neighbor instances has been proposed as an alternative option \citep{zhu-etal-2008-active,zhu2009active}.


\subsubsection{Discriminative\footnote{Some works also use the word ``diversity,'' however we specifically preserve this word for batch-diversity in \S\ref{sec:c5a:Qdiv}.}}
\label{sec:c5a:Qdis}

Another direction is to select instances that are \textbf{different from already labeled instances}.
Again, for NLP tasks, simple feature-based metrics can be utilized for this purpose by preferring instances with more unseen $n$-grams or out-of-vocabulary words \citep{eck-etal-2005-low,bloodgood-callison-burch-2010-bucking,erdmann-etal-2019-practical}.
Generally, similarity scores can also be utilized to select the instances that are less similar to the labeled set \citep{kim-etal-2006-mmr,zhang2018active,zeng-etal-2019-empirical}.
Another interesting idea is to train a model to discriminate the labeled and unlabeled sets. \citet{gissin2019discriminative} directly train a classifier for this purpose, while naturally adversarial training can be also adopted \citep{sinha2019variational,deng2018adversarial}.
In domain adaptation scenarios, the same motivation leads to the usage of a domain separator to filter instances \citep{rai-etal-2010-domain}.


\subsubsection{Batch Diversity}
\label{sec:c5a:Qdiv}

Ideally, only one most useful instance would be selected in each iteration. However, it is more efficient and practical to adopt \textbf{batch-mode AL} \citep{settles2009active}, where each time a batch of instances is selected. In this case, we need to consider the dissimilarities not only between selected instances and labeled ones but also within the selected batch.

To select a batch of diverse instances, there are two common approaches. 1) \textbf{Iterative selection} collects the batch in an iterative greedy way \citep{brinker2003incorporating,shen-etal-2004-multi}. In each iteration, an instance is selected by comparing it with previously chosen instances to avoid redundancy. 
Some more advanced diversity-based criteria, like coreset \citep{geifman2017deep,sener2018active} and determinantal point processes \citep{shi-etal-2021-diversity}, can also be approximated in a similar way. 2) \textbf{Clustering-based} methods partition the unlabeled data into clusters and select instances among them \citep{tang-etal-2002-active,xu2003representative,shen-etal-2004-multi,nguyen2004active,zhdanov2019diverse,yu-etal-2022-actune,maekawa-etal-2022-low}. Since the chosen instances come from different clusters, diversity can be achieved to some extent.

To calculate similarity, in addition to comparing the input features or intermediate representations, other methods are also investigated, such as utilizing model-based similarity \citep{hazra-etal-2021-active2}, gradients \citep{Ash2020Deep,kim-2020-deep}, and masked LM surprisal embeddings \citep{yuan-etal-2020-cold}.


\subsection{Hybrid}
\label{sec:c5a:Qh}


There is no surprise that informativeness and representativeness can be combined for instance querying, leading to hybrid strategies.
A \textbf{simple combination} can be used to merge multiple criteria into one. This can be achieved by a weighted sum \citep{kim-etal-2006-mmr,chen-etal-2011-enhancing} or multiplication \citep{settles-craven-2008-analysis,zhu-etal-2008-active}.

There are several strategies to \textbf{naturally integrate} multiple criteria. Examples include (uncertainty) weighted clustering \citep{zhdanov2019diverse}, diverse gradient selection \citep{Ash2020Deep,kim-2020-deep} where the gradients themselves contain uncertainty information (\S\ref{sec:c5a:Qgrad}) and determinantal point processes (DPP) with quality-diversity decomposition \citep{shi-etal-2021-diversity}.

Moreover, \textbf{multi-step querying}, which applies multiple criteria in series, is another natural hybrid method. For example, one can consider first filtering certain highly uncertain instances and then performing clustering to select a diverse batch from them \citep{xu2003representative,shen-etal-2004-multi,mirroshandel-etal-2011-active}. An alternative strategy of selecting the most uncertain instances per cluster has also been utilized \citep{tang-etal-2002-active}.

Instead of statically merging into one query strategy, \textbf{dynamic combination} may better fit the AL learning process, since different strategies may excel at different AL phases. For example, at the start of AL, uncertainty sampling may be unreliable due to little labeled data, and representativeness-based methods could be preferable, whereas in later stages where we have enough data and target finer-grained decision boundaries, uncertainty may be a suitable strategy. DUAL \citep{donmez2007dual} is such a dynamic strategy that can switch from a density-based selector to an uncertainty-based one. \citet{ambati-etal-2011-multi} further propose GraDUAL, which gradually switches strategies within a switching range. \citet{wu-etal-2017-active} adopt a similar idea with a pre-defined monotonic function to control the combination weights.



\section{Query and Annotation}
\label{sec:c5a:A}


\subsection{AL for Structured Prediction}
\label{sec:c5a:sp}

AL has been widely studied for classification tasks, while in NLP, many tasks involve \emph{structured prediction}. In these tasks, the system needs to output a structured object consisting of a group of inter-dependent variables \citep{smith2011linguistic}, such as a label sequence or a parse tree. Special care needs to be taken when querying and annotating for these more complex tasks \citep{thompson1999active}. One main decision is whether to annotate full structures for input instances (\S\ref{sec:c5a:full}), or allow the annotation of only partial structures (\S\ref{sec:c5a:partial}).


\subsubsection{Full-structure AL}
\label{sec:c5a:full}

First, if we regard the full output structure of an instance as a whole and perform query and annotation at the full-instance level, then AL for structured prediction tasks is not very different than for simpler classification tasks. Nevertheless, considering that the output space is usually exponentially large and infeasible to explicitly enumerate, querying may require further inspection.

Some \textbf{uncertainty sampling} strategies, such as entropy, need to consider the full output space. Instead of the infeasible explicit enumeration, dynamic-programming algorithms that are similar to the ones in decoding and inference processes can be utilized, such as algorithms for tree-entropy \citep{hwa-2000-sample,hwa-2004-sample} and sequence-entropy \citep{mann-mccallum-2007-efficient,settles-craven-2008-analysis}.

Instead of considering the full output space, \textbf{top-$k$ approximation} is a simpler alternative that takes $k$-best predicted structures as a proxy. This is also a frequently utilized method \citep{tang-etal-2002-active,kim-etal-2006-mmr,rocha-sanchez-2013-towards}.

For disagreement-based strategies, the measurement of \textbf{partial disagreement} may be required, since full-match can be too strict for structured objects.
Fine-grained evaluation scores can be reasonable choices for this purpose, such as F1 score for sequence labeling \citep{ngai-yarowsky-2000-rule}.

Since longer instances usually have larger uncertainties and might be preferred, \textbf{length normalization} is a commonly-used heuristic to avoid this bias \citep{tang-etal-2002-active,hwa-2000-sample,hwa-2004-sample,shen2018deep}. Yet, \citet{settles-craven-2008-analysis} argue that longer sequences should not be discouraged and may contain more information.

Instead of directly specifying the full utility of an instance, \textbf{aggregation} is also often utilized by gathering utilities of its sub-structures, usually along the factorization of the structured modeling. For example, the sequence uncertainty can be obtained by summing or averaging the uncertainties of all the tokens \citep{settles-craven-2008-analysis}. Other aggregation methods are also applicable, such as weighted sum by word frequency \citep{ringger-etal-2007-active} or using only the most uncertain (least probable) one \citep{myers-palmer-2021-tuning,liu2022ltp}.


\subsubsection{Partial-structure AL}
\label{sec:c5a:partial}

A structured object can be decomposed into smaller sub-structures with different training utilities. For example, in a dependency tree, functional relations are usually easier to judge while prepositional attachment links may be more informative for the learning purpose. This naturally leads to AL with partial structures, where querying and annotating can be performed at the sub-structure level.

Factorizing full structures into the \textbf{finest-grained sub-structures} and regarding them as the annotation units could be a natural choice. Typical examples include individual tokens for sequence labeling \citep{marcheggiani-artieres-2014-experimental}, word boundaries for segmentation \citep{neubig-etal-2011-pointwise,li-etal-2012-active}, syntactic-unit pairs for dependency parsing \citep{sassano-kurohashi-2010-using} and mention pairs for coreference \citep{gasperin-2009-active,miller-etal-2012-active,sachan2015active}. The querying strategy for the sub-structures can be similar to the classification cases, though inferences are usually needed to calculate marginal probabilities. Moreover, if full structures are desired as annotation outputs, semi-supervised techniques such as self-training (\S\ref{sec:c5a:learn}) could be utilized to assign pseudo labels to the unannotated parts \citep{tomanek-hahn-2009-semi,majidi-crane-2013-active}.

At many times, choosing \textbf{larger sub-structures} is preferable, since partial annotation still needs the understanding of larger contexts and frequently jumping among different contexts may require more reading time (\S\ref{sec:c5a:measure}). Moreover, increasing the sampling granularity may mitigate the missed class effect, where certain classes may be overlooked \citep{tomanek-etal-2009-proper}. Typical examples of larger sub-structures include sub-sequences for sequence labeling \citep{shen-etal-2004-multi,chaudhary-etal-2019-little,radmard-etal-2021-subsequence}, word-wise head edges for dependency parsing \citep{flannery-mori-2015-combining,li-etal-2016-active}, neighborhood pools \citep{laws-etal-2012-active} or mention-wise anaphoric links \citep{li-etal-2020-active-learning,espeland-etal-2020-enhanced} for coreference, and phrases for MT \citep{bloodgood-callison-burch-2010-bucking,miura-etal-2016-selecting,hu-neubig-2021-phrase}. In addition to increasing granularity, \textbf{grouping queries} can also help to make annotation easier, such as adopting a two-stage selection of choosing uncertain tokens from uncertain sentences \citep{mirroshandel-nasr-2011-active,flannery-mori-2015-combining} and selecting nearby instances in a row \citep{miller-etal-2012-active}.

For AL with partial structures, \textbf{output modeling} is of particular interest since the model needs to learn from partial annotations. If directly using local discriminative models where each sub-structure is decided independently, learning with partial annotations is straightforward since the annotations are already complete to the models \citep{neubig-etal-2011-pointwise,flannery-mori-2015-combining}. For more complex models that consider interactions among output sub-structures, such as global models, special algorithms are required to learn from incomplete annotations \citep{scheffer2001active,wanvarie2011active,marcheggiani-artieres-2014-experimental,li-etal-2016-active}.
One advantage of these more complex models is the interaction of the partial labels and the remaining parts. 
For example, considering the \textbf{output constraints} for structured prediction tasks, combining the annotated parts and the constraints may reduce the output space of other parts and thus lower their uncertainties, leading to better queries \citep{roth2006margin,sassano-kurohashi-2010-using,mirroshandel-nasr-2011-active}. 
More generally, the annotation of one label can intermediately influence others with cheap re-inference, which can help batch-mode selection \citep{marcheggiani-artieres-2014-experimental} and interactive correction \citep{culotta2005reducing}.

In addition to classical structured-prediction tasks, classification tasks can also be cast as structured predictions with partial labeling.
\textbf{Partial feedback} is an example that is adopted to make the annotating of classification tasks simpler, especially when there are a large number of target labels. For example, annotators may find it much easier to answer yes/no questions \citep{hu2018active} or rule out negative classes \citep{lippincott-van-durme-2021-active} than to identify the correct one.



\subsection{Annotation Cost}
\label{sec:c5a:cost}

AL mainly aims to reduce real annotation cost and we discuss several important topics on this point.

\subsubsection{Cost Measurement}
\label{sec:c5a:measure}

Most AL works adopt simple measurements of unit cost, that is, assuming that annotating each instance requires the same cost. Nevertheless, the annotation efforts for different instances may vary \citep{settles2008active}. For example, longer sentences may cost more to annotate than shorter ones. Because of this, many works assume unit costs to tokens instead of sequences, which may still be inaccurate. Especially, AL tends to select difficult and ambiguous instances, which may require more annotation efforts \citep{hachey-etal-2005-investigating,lynn-etal-2012-active}. It is important to \textbf{properly measure annotation cost} since the measurement directly affects the evaluation of AL algorithms. The comparisons of query strategies may vary if adopting different cost measurement \citep{haertel-etal-2008-assessing,bloodgood-callison-burch-2010-bucking,chen2015study}.

Probably the best cost measurement is the actual \textbf{annotation time} \citep{baldridge-palmer-2009-well}. Especially, when the cost comparisons are not that straightforward, such as comparing annotating data against writing rules \citep{ngai-yarowsky-2000-rule} or partial against full annotations \citep[\S\ref{sec:c5a:sp};][]{flannery-mori-2015-combining,li-etal-2016-active,li-etal-2020-active-learning}, time-based evaluation is an ideal choice. This requires actual annotating exercises rather than simulations.

Since cost measurement can also be used for querying (\S\ref{sec:c5a:sensitive}), it would be helpful to be able to \textbf{predict the real cost} before annotating. This can be cast as a regression problem, for which several works learn a linear cost model based on input features \citep{settles2008active,ringger-etal-2008-assessing,haertel-etal-2008-assessing,arora-etal-2009-estimating}.


\subsubsection{Cost-sensitive Querying}
\label{sec:c5a:sensitive}

Given the goal of reducing actual cost, the querying strategies should also take it into consideration. That is, we want to select not only high-utility instances but also low-cost ones. A natural cost-sensitive querying strategy is \textbf{return-on-investment} \citep[ROI;][]{haertel2008return,settles2008active,donmez2008proactive}. In this strategy, instances with higher net benefit per unit cost are preferred, which is equivalent to dividing the original querying utility by cost measure. \citet{tomanek-hahn-2010-comparison} evaluate the effectiveness of ROI together with two other strategies, including constraining maximal cost budget per instance and weighted rank combination. \citet{haertel-etal-2015-analytic} provide further analytic and empirical evaluation, showing that ROI can reduce total cost.

In real AL scenarios, things can be much more complex. For example, there can be multiple annotators with different expertise \citep{baldridge-palmer-2009-well,huang2017cost,cai2020cost}, and the annotators may refuse to answer or make mistakes \citep{donmez2008proactive}. Being aware of these scenarios, \citet{donmez2008proactive} propose \textbf{proactive learning} to jointly select the optimal oracle and instance. \citet{li-etal-2017-proactive} further extend proactive learning to NER tasks.


\subsubsection{Directly Reducing Cost}
\label{sec:c5a:cut}

In addition to better query strategies, there are other ways of directly reducing annotation cost, such as computer-assisted annotation. In AL, models and annotators usually interact in an indirect way where models only query the instances to present to the annotators, while there could be closer interactions. 

\textbf{Pre-annotation} is such an idea, where not only the raw data instances but also the model's best or top-$k$ predictions are sent to the annotators to help them make decisions. If the model's predictions are reasonable, the annotators can simply select or make a few corrections to obtain the gold annotations rather than creating from scratch. This method has been shown effective when combined with AL \citep{baldridge-osborne-2004-active,vlachos-2006-active,ringger-etal-2008-assessing,skeppstedt-2013-annotating,canizares-diaz-etal-2021-active}. Post-editing for MT is also a typical example \citep{dara-etal-2014-active}.

Moreover, the models could provide help at \textbf{real annotating time}. For example, \citet{culotta2005reducing} present an interactive AL system where the user's corrections can propagate to the model, which generates new predictions for the user to further refine. Interactive machine translation (IMT) adopts a similar idea, where the annotator corrects the first erroneous character, based on which the model reproduces the prediction. AL has also been combined with IMT to further reduce manual efforts \citep{gonzalez-rubio-etal-2012-active,peris-casacuberta-2018-active,gupta-etal-2021-investigating}.

Crowdsourcing is another way to reduce annotation costs and can be combined with AL. We provide more discussions on this in \S\ref{sec:c5a:O}.


\subsubsection{Wait Time}
\label{sec:c5a:wait}

In AL iterations, the annotators may need to wait for the training and querying steps (Line 7 and 4 in Algorithm~\ref{algo:c5a:al}). This wait time may bring some hidden costs, thus more efficient querying and training would be preferable for faster turnarounds.

To speed up \textbf{querying}, sub-sampling is a simple method to deal with large unlabeled pools \citep{roy2001toward,ertekin2007learning,tsvigun-etal-2022-towards,maekawa-etal-2022-low}. For some querying strategies, pre-calculating and caching unchanging information can also help to speed up \citep{ashrafi-asli-etal-2020-optimizing,citovsky2021batch}. In addition, approximation with $k$-nearest neighbours can also be utilized to calculate density \citep{zhu2009active} or search for instances after adversarial attacks \citep{ru-etal-2020-active}.

To reduce \textbf{training} time, a seemingly reasonable strategy is to apply incremental training across AL iterations, that is, continuing training previous models on the new instances. However, \citet{ash2020warm} show that this type of warm-start may lead to sub-optimal performance for neural models and many recent AL works usually train models from scratch \citep{hu2018active,ein-dor-etal-2020-active}. Another method is to use an efficient model for querying and a more powerful model for final training. However, this might lead to sub-optimal results, which will be discussed in \S\ref{sec:c5a:mismatch}.

Another idea to reduce wait time is to simply allow querying with \textbf{stale information}. Actually, batch-mode AL (\S\ref{sec:c5a:Qdiv}) is such an example where instances in the same batch are queried with the same model. \citet{haertel-etal-2010-parallel} propose parallel AL, which maintains separate loops of annotating, training, and scoring, and allows dynamic and parameterless instance selection at any time.


\section{Model and Learning}
\label{sec:c5a:M}



\subsection{Model Mismatch}
\label{sec:c5a:mismatch}

While it is natural to adopt the same best-performing model throughout the AL process, there are cases where the query and final (successor) models can mismatch \citep{lewis1994heterogeneous}. Firstly, more efficient models are preferable for querying to reduce wait time (\S\ref{sec:c5a:wait}). Moreover, since data usually outlive models, re-using AL-base data to train another model would be desired \citep{baldridge-osborne-2004-active,tomanek-etal-2007-approach}.
Several works show that model mismatch may make the gains from AL be negligible or even negative \citep{baldridge-osborne-2004-active,lowell-etal-2019-practical,shelmanov-etal-2021-active}, which raises concerns about the utilization of AL in practice. 

For efficiency purposes, distillation can be utilized to improve querying efficiency while keeping reasonable AL performance. \citet{shelmanov-etal-2021-active} show that using a smaller distilled version of a pre-trained model for querying does not lead to too much performance drop. \citet{tsvigun-etal-2022-towards} combine this idea with pseudo-labeling and sub-sampling to further reduce computational cost. Similarly, \citet{nguyen-etal-2022-famie} keep a smaller proxy model for query and synchronize the proxy with the main model by distillation.


\subsection{Learning}
\label{sec:c5a:learn}

AL can be combined with other advanced learning techniques to further reduce required annotations.

\paragraph{Semi-supervised learning.} Since AL usually assumes an unlabeled pool, semi-supervised learning can be a natural fit. Combining these two is not a new idea: \citep{mccallum1998employing} adopt the EM algorithm to estimate the outputs of unlabeled data and utilize them for learning. This type of self-training or pseudo-labeling technique is often utilized in AL \citep{tomanek-hahn-2009-semi,majidi-crane-2013-active,yu-etal-2022-actune}. With a similar motivation, \citep{dasgupta-ng-2009-mine} use an unsupervised algorithm to identify the unambiguous instances to train an active learner. For the task of word alignment, which can be learned in an unsupervised manner, incorporating supervision with AL can bring further improvements in a data-efficient way \citep{ambati-etal-2010-active-learning,ambati-etal-2010-active-semi}.


\paragraph{Transfer learning.}
AL can be easily combined with transfer learning, another technique to reduce required annotations. Utilizing pre-trained models is already a good example \citep{ein-dor-etal-2020-active,yuan-etal-2020-cold,tamkin2022active} and continual training \citep{gururangan-etal-2020-dont} can also be applied \citep{hua-wang-2022-efficient,margatina-etal-2022-importance}. Moreover, transductive learning is commonly combined with AL by transferring learning signals from different domains \citep{chan-ng-2007-domain,shi2008actively,rai-etal-2010-domain,saha2011active,wu-etal-2017-active,kasai-etal-2019-low,yuan-etal-2022-adapting} or languages \citep{qian-etal-2014-bilingual,fang-cohn-2017-model,fang-etal-2017-learning,chaudhary-etal-2019-little,chaudhary-etal-2021-reducing,moniz-etal-2022-efficiently}. Initializing the model via meta-learning has also been found helpful \citep{zhu-etal-2022-shot}. In addition to the task model, the model-based query policy (\S\ref{sec:c5a:Qpred}) is also often obtained with transfer learning.


\paragraph{Weak supervision.} AL can also be combined with weakly supervised learning. Examples include learning from execution results for semantic parsing \citep{ni2020merging}, labeling based on structure vectors for entity representations \citep{qian-etal-2020-learning}, learning from gazetteers for sequence labeling \citep{brantley-etal-2020-active} and interactively discovering labeling rules \citep{zhang-etal-2022-prompt}.


\paragraph{Data augmentation.} Augmentation is also applicable in AL and has been explored with iterative back-translation \citep{zhao-etal-2020-active}, mixup for sequence labeling \citep{zhang-etal-2020-seqmix} and phrase-to-sentence augmentation for MT \citep{hu-neubig-2021-phrase}. As discussed in \S\ref{sec:c5a:Quncer}, augmentation can also be helpful for instance querying \citep{jiang-etal-2020-camouflaged,zhang-etal-2022-allsh}.
Another interesting scenario involving augmentation and AL is query synthesis, which directly generates instances to be annotated instead of selecting existing unlabeled ones. Though synthesizing texts is still a hard problem generally, there have been successful applications for simple classification tasks \citep{schumann-rehbein-2019-active,quteineh-etal-2020-textual}.


\section{Starting and Stopping AL}
\label{sec:c5a:S}


\subsection{Starting AL}
\label{sec:c5a:start}

While there are cases where there are already enough labeled data to train a reasonable model and AL is utilized to provide further improvements \citep{bloodgood-callison-burch-2010-bucking,geifman2017deep}, at many times we are facing the cold-start problem, where instances need to be selected without a reasonable model. Especially, how to select the seed data to start the AL process is an interesting question, which may greatly influence the performance in initial AL stages \citep{tomanek-etal-2009-proper,horbach-palmer-2016-investigating}.

Random sampling is probably the most commonly utilized strategy, which is reasonable since it preserves the original data distribution. 
Some representativeness-based querying strategies (\S\ref{sec:c5a:Qr}) can also be utilized, for example, selecting points near the clustering centroids is a way to obtain representative and diverse seeds \citep{kang2004using,zhu-etal-2008-active,hu2010off}. 
Moreover, some advanced learning techniques (\S\ref{sec:c5a:learn}) can also be helpful here, such as transfer learning \citep{wu-etal-2017-active} and unsupervised methods \citep{vlachos-2006-active,dasgupta-ng-2009-mine}.
In addition, language model can be a useful tool, with which \citet{dligach-palmer-2011-good} select low-probability words in the context of word sense disambiguation and \citet{yuan-etal-2020-cold} choose cluster centers with surprisal embeddings by pre-trained contextualized LMs.


\subsection{Stopping AL}
\label{sec:c5a:stop}

When adopting AL in practice, it would be desirable to know the time to stop AL when the model performance is already near the upper limits, before running out of all the budgets. 
For this purpose, a stopping criterion is needed, which checks certain metrics satisfying certain conditions.
There can be simple heuristics. For example, AL can be stopped when all unlabeled instances are no closer than any of the support vectors with an SVM \citep{schohn2000less,ertekin2007learning} or no new $n$-grams remain in the unlabeled set for MT \citep{bloodgood-callison-burch-2010-bucking}. Nevertheless, these are not generic solutions.
For the design of a general stopping criterion, there are three main aspects to consider: \emph{metric}, \emph{dataset} and \emph{condition}.

For the \textbf{metric}, measuring performance on a development set seems a natural option. However, the results would be unstable if this set is too small and it would be impractical to assume a large development set. Cross-validation on the training set also has problems since the labeled data by AL is usually biased.
In this case, metrics from the query strategies can be utilized. Examples include uncertainty or confidence \citep{zhu-hovy-2007-active,vlachos2008stopping}, disagreement \citep{tomanek-etal-2007-approach,tomanek-hahn-2008-approximating,olsson-tomanek-2009-intrinsic}, estimated performance \citep{laws-schutze-2008-stopping}, expected error \citep{zhu-etal-2008-learning}, confidence variation \citep{ghayoomi-2010-using}, as well as actual performance on the selected instances \citep{zhu-hovy-2007-active}. Moreover, comparing the predictions between consecutive AL iterations is another reasonable option \citep{zhu-etal-2008-multi,bloodgood-vijay-shanker-2009-method}.

The \textbf{dataset} to calculate the stopping metric requires careful choosing. The results could be unstable if not adopting a proper set \citep{tomanek-hahn-2008-approximating}. Many works suggest that a separate unlabeled dataset should be utilized \citep{tomanek-hahn-2008-approximating,vlachos2008stopping,bloodgood-vijay-shanker-2009-method,beatty2019use,kurlandski2022impact}. Since the stopping metrics usually do not rely on gold labels, this dataset could potentially be very large to provide more stable results, though wait time would be another factor to consider in this case (\S\ref{sec:c5a:wait}).

The \textbf{condition} to stop AL is usually comparing the metrics to a pre-defined threshold. Earlier works only look at the metric at the current iteration, for example, stopping if the uncertainty or the error is less than the threshold \citep{zhu-hovy-2007-active}. In this case, the threshold is hard to specify since it relies on the model and the task. \citep{zhu-etal-2008-multi} cascade multiple stopping criteria to mitigate this reliance. A more stable option is to track the change of the metrics over several AL iterations, such as stopping when the confidence consistently drops \citep{vlachos2008stopping}, the changing rate flattens \citep{laws-schutze-2008-stopping} or the predictions stabilize across iterations \citep{bloodgood-vijay-shanker-2009-method,bloodgood-grothendieck-2013-analysis}.

\citet{pullar2021hitting} provide an empirical comparison over common stopping criteria and would be a nice reference. Moreover, stopping AL can be closely related to performance prediction and early stopping. Especially, the latter can be of particular interest since learning in early AL stages is a low-resource problem and how to perform early stopping may also require careful considerations.


\section{Other Aspects}
\label{sec:c5a:O}

We describe some other aspects that are frequently seen when applying AL to NLP.

\paragraph{Crowdsourcing and Noise.} Crowdsourcing is another way to reduce annotation costs by including non-expert annotations \citep{snow-etal-2008-cheap}. Naturally, AL and crowdsourcing may also be combined with the hope to further reduce cost \citep{ambati-etal-2010-active,laws-etal-2011-active,yan2011active,fang2014active,zhao2020integrating}. One specific factor to consider in this case is the noises in the crowdsourced data, since noisy data may have a negative impact on the effectiveness of AL \citep{rehbein-ruppenhofer-2011-evaluating}. Cost-sensitive querying strategies (\S\ref{sec:c5a:sensitive}) can be utilized to select both annotators and instances by estimating labelers' reliability \citep{yan2011active,fang2014active}. Requiring multiple annotations per instance and then consolidating is also applicable \citep{laws-etal-2011-active}. \citet{lin-etal-2019-alpacatag} provide a framework that enables automatic crowd consolidation for AL on the tasks of sequence labeling.


\paragraph{Multiple Targets.} In many cases, we may want to consider multiple targets rather than only one, for example, annotating instances in multiple domains \citep{xiao-guo-2013-online,he2021multi,longpre2022active} or multiple languages \citep{haffari-sarkar-2009-active,qian-etal-2014-bilingual,moniz-etal-2022-efficiently}. Moreover, there may be multiple target tasks, where multi-task learning (MTL) can interact with AL \citep{reichart-etal-2008-multi,ambati-etal-2011-active,rocha-sanchez-2013-towards,ikhwantri-etal-2018-multi,zhu-etal-2020-multitask,rotman-reichart-2022-multi}. In these scenarios with multiple targets, naturally, strategies that consider all the targets are usually more preferable. \citet{reichart-etal-2008-multi} show that a query strategy that considers all target tasks obtains the overall best performance for MTL. \citet{moniz-etal-2022-efficiently} suggest that joint learning across multiple languages using a single model outperforms other strategies such as equally dividing budgets or allocating only for a high-resource language and then performing the transfer.


\paragraph{Data Imbalance.} Imbalance is a frequently occurring phenomenon in NLP and AL can have interesting interactions with it. On the one hand, as in plain learning scenarios, AL should take data imbalance into considerations, with modifications to the model \citep{bloodgood-vijay-shanker-2009-taking}, learning algorithm \citep{zhu-hovy-2007-active} and query strategies \citep{tomanek-etal-2009-proper,escudeiro-jorge-2010-confidence,li-etal-2012-active-learning}. On the other hand, AL can be utilized to address the data imbalance problem and build better data \citep{ertekin2007learning,tomanek2009reducing,attenberg2013class,mottaghi2020medical,mussmann-etal-2020-importance}.



\section{Related Topics and Future Directions}
\label{sec:c5a:dir}


\subsection{Related Topics}

There are many related topics that could be explored together with AL. Other data-efficient learning methods such as semi-supervised and transfer learning are naturally compatible with AL (\S\ref{sec:c5a:learn}). Curriculum learning \citep{bengio2009curriculum}, which arranges training instances in a meaningful order, may also be integrated with AL \citep{platanios-etal-2019-competence,zhao2020reinforced,jafarpour-etal-2021-active}. Uncertainty \citep{gawlikowski2021survey}, outlier detection \citep{hodge2004survey} and performance prediction \citep{xia-etal-2020-predicting} can be related to instance querying. 
Crowdsourcing can be adopted to further reduce annotation cost (\S\ref{sec:c5a:O}). Model efficiency \citep{menghani2021efficient} would be crucial to reduce wait time (\S\ref{sec:c5a:wait}). AL is a typical type of human-in-the-loop framework \citep{wang-etal-2021-putting}, and it will be interesting to explore more human-computer interaction techniques in AL.

\subsection{Future Directions}

\paragraph{Complex tasks.} AL is mostly adopted for simple classification, while there are many more complex tasks in NLP. For example, except for MT, generation tasks have been much less thoroughly explored with AL. Tasks with more complex inputs such as NLI and QA also require extra care when using AL; obtaining unlabeled data is already non-trivial. Nevertheless, preliminary work has shown that AL can be helpful for data collection for such tasks \citep{mussmann-etal-2020-importance}.

\paragraph{Beyond direct target labeling.} In addition to directly annotating target labels, AL can also be utilized in other ways to help the target task, such as labeling features or rationales \citep{melville-sindhwani-2009-active,druck-etal-2009-active,sharma-etal-2015-active}, annotating explanations \citep{liang-etal-2020-alice}, evaluation \citep{mohankumar-khapra-2022-active} and rule discovery \citep{zhang-etal-2022-prompt}.

\paragraph{AL in practice.} Most AL works simulate annotations on an existing labeled dataset. Though this method is convenient for algorithm development, it ignores several challenges of applying AL in practice. As discussed in this survey, real annotation cost (\S\ref{sec:c5a:measure}), efficiency and wait time (\S\ref{sec:c5a:wait}), data reuse (\S\ref{sec:c5a:mismatch}) and starting and stopping (\S\ref{sec:c5a:S}) are all important practical aspects which may not emerge in simulation. Moreover, since the AL process usually cannot be repeated multiple times, how to select the query strategy and other hyper-parameters remains a great challenge. It will be critical to address these issues to bring AL into practical use \citep{rehbein-etal-2010-bringing,attenberg2011inactive,settles2011theories,lowell-etal-2019-practical} and make it more widely utilized \citep{tomanek-olsson-2009-web}.


\section*{Limitations}

There are several limitations of this work. First, we mainly focus on AL works in the context of NLP, while AL works in other fields may also present ideas that could be utilized for NLP tasks. For example, many querying strategies originally developed with CV tasks could be naturally adopted to applications in NLP \citep{ren2021survey}. We encourage the readers to refer to other surveys mentioned in \S\ref{sec:c5a:intro} for additional related AL works. Moreover, the descriptions in this survey are mostly brief in order to provide a more comprehensive coverage within page limits. We mainly present the works in meaningful structured groups rather than plainly describing them in unstructured sequences, and we hope that this work can serve as an index where more details can be found in corresponding works. Finally, this is a pure survey without any experiments or empirical results. It would be helpful to perform comparative experiments over different AL strategies, which could provide more meaningful guidance \citep{zhan2022comparative}. We leave this to future work.

\bibliography{al4nlp}
\bibliographystyle{acl_natbib}

\appendix


\section{Tasks}
\label{sec:c5a:T}

In this section, we list representative works for different NLP tasks. According to the output structures, the tasks are further categorized into four groups: classification, sequence labeling, complex structured prediction, and generation.

\paragraph{Classification} denotes the tasks whose output consists of only one variable. Text classification that assigns a target label to an input text sequence is a typical example. Pairwise classification and word-level classification are also commonly seen in NLP.
\begin{itemize}[noitemsep,leftmargin=*]
\item \textbf{Text classification:} Please refer to the paper table mentioned in (\S\ref{sec:c5a:pt}) for related works. We do not list them here since there are too many.
\item \textbf{Pairwise classification:} \citep{griesshaber-etal-2020-fine,bai-etal-2020-pre,mussmann-etal-2020-importance}
\item \textbf{Word sense disambiguation (WSD):} \citep{fujii-etal-1998-selective,chen-etal-2006-empirical,chan-ng-2007-domain,zhu-hovy-2007-active,zhu-etal-2008-active,imamura-etal-2009-combination,martinez-alonso-etal-2015-active}
\end{itemize}

\paragraph{Sequence labeling} is probably the most commonly seen structured prediction task in NLP. It aims to predict a sequence of labels, among which there may be interactions and constraints.
\begin{itemize}[noitemsep,leftmargin=*]
\item \textbf{Part-of-speech (POS):} \citep{engelson-dagan-1996-minimizing,ringger-etal-2007-active,haertel-etal-2008-assessing,marcheggiani-artieres-2014-experimental,fang-cohn-2017-model,brantley-etal-2020-active,chaudhary-etal-2021-reducing}
\item \textbf{(Named) entity recognition (NER/ER):} \citep{shen-etal-2004-multi,culotta2005reducing,kim-etal-2006-mmr,settles-craven-2008-analysis,tomanek-hahn-2009-semi,marcheggiani-artieres-2014-experimental,chen2015study,li-etal-2017-proactive,shen2018deep,siddhant-lipton-2018-deep,erdmann-etal-2019-practical,chaudhary-etal-2019-little,chang2020using,brantley-etal-2020-active,hazra-etal-2021-active2,shelmanov-etal-2021-active,radmard-etal-2021-subsequence,dossou-etal-2022-afrolm}
\item \textbf{Segmentation:} \citep{ngai-yarowsky-2000-rule,sassano-2002-empirical,neubig-etal-2011-pointwise,li-etal-2012-active,marcheggiani-artieres-2014-experimental,cai2021ne}
\item \textbf{Natural language understanding (NLU):} \citep{hadian-sameti-2014-active,deng2018adversarial,peshterliev-etal-2019-active,zhu-etal-2020-multitask}
\end{itemize}

\paragraph{Complex structure prediction} in this work denotes the structure prediction tasks that are more complex than sequence labeling, and have \emph{explicit} connections (alignments) between inputs and outputs. They usually aim to extract relational structures among input elements.
\begin{itemize}[noitemsep,leftmargin=*]
\item \textbf{Parsing:} \citep{hwa-2000-sample,tang-etal-2002-active,baldridge-osborne-2003-active,baldridge-osborne-2004-active,hwa-2004-sample,reichart-rappoport-2009-sample,sassano-kurohashi-2010-using,atserias-etal-2010-active,mirroshandel-nasr-2011-active,majidi-crane-2013-active,flannery-mori-2015-combining,li-etal-2016-active,shi-etal-2021-diversity}
\item \textbf{Semantic role labeling (SRL):} \citep{roth2006margin,wang2017active,ikhwantri-etal-2018-multi,siddhant-lipton-2018-deep,koshorek-etal-2019-limits,myers-palmer-2021-tuning}
\item \textbf{Coreference:} \citep{gasperin-2009-active,miller-etal-2012-active,laws-etal-2012-active,zhao-ng-2014-domain,sachan2015active,li-etal-2020-active-learning,espeland-etal-2020-enhanced,yuan-etal-2022-adapting}
\item \textbf{Relation-related:} \citep{roth2008active,bloodgood-vijay-shanker-2009-taking,mirroshandel-etal-2011-active,fu-grishman-2013-efficient,canizares-diaz-etal-2021-active,mallart-etal-2021-active,seo2022active,zhang-etal-2022-prompt}
\item \textbf{Event-related:} \citep{cao-etal-2015-improving,shen2021active,lee-etal-2022-crudeoilnews}
\item \textbf{Word alignment:} \citep{ambati-etal-2010-active-learning,ambati-etal-2010-active-semi,rocha-sanchez-2013-towards}
\item \textbf{Entity alignment/resolution:} \citep{kasai-etal-2019-low,liu-etal-2021-activeea}
\end{itemize}

\paragraph{Generation} refers to the tasks that aim to generate a sequence of tokens. We differentiate them from plain structured prediction tasks since there are usually no explicit alignments between input and output sub-parts in the supervision and such alignments are usually implicitly modeled, especially in recent sequence-to-sequence neural models. MT is a typical generation task, where we further separate traditional statistical machine translation (SMT) and recent neural machine translation (NMT). We also include semantic parsing here, since recent works usually cast it as a sequence-to-sequence generation task.
\begin{itemize}[noitemsep,leftmargin=*]
\item \textbf{SMT:} \citep{eck-etal-2005-low,haffari-etal-2009-active,haffari-sarkar-2009-active,ananthakrishnan-etal-2010-semi,bloodgood-callison-burch-2010-bucking,ambati-etal-2010-active,ananthakrishnan-etal-2010-discriminative,gonzalez-rubio-etal-2012-active,rocha-sanchez-2013-towards,logacheva-specia-2014-confidence,logacheva-specia-2014-quality,miura-etal-2016-selecting}
\item \textbf{NMT:} \citep{peris-casacuberta-2018-active,liu-etal-2018-learning,zhang2018active,zeng-etal-2019-empirical,zhao-etal-2020-active,hu-neubig-2021-phrase,gupta-etal-2021-investigating,zhou-waibel-2021-active,hazra-etal-2021-active2,mendoncca2022onception}
\item \textbf{Semantic parsing:} \citep{duong-etal-2018-active,ni2020merging,sen-yilmaz-2020-uncertainty}
\item \textbf{Others:} \citep{mairesse-etal-2010-phrase,deng2018adversarial,tsvigun-etal-2022-active}
\end{itemize}


\section{Surveying Process}
\label{sec:c5a:pt}

In this section, we provide more details of our surveying process:
\begin{itemize}
    \item For the ACL Anthology, we search for papers with the keyword ``active'' in titles (by grepping the ``Full Anthology BibTeX file''\footnote{\url{https://aclanthology.org/anthology.bib.gz}}). There can be related papers that are missed from this simple keyword search, but as we read along the filtered list, we gradually include the notable missing ones.
    \item We also include papers outside the ACL Anthology. First, we look for papers by searching with the key phrase ``active learning'' on Arxiv (in the field of cs.CL, excluding those already appearing in ACL Anthology). Moreover, we also collect related works in other venues, such as AI/ML conferences and journals. For these venues, we do (can) not perform extensive searches due to high volume (and that many are unrelated to our focus on NLP). We mainly collect related papers in these adjacent venues by following the references from the papers already surveyed.
\end{itemize}
We also create a table for the related papers (with detailed categorizations), which can be found at this link: \url{https://github.com/zzsfornlp/zmsp/blob/main/msp2/docs/al4nlp/readme.md}.

\end{document}